\documentclass{article}

\usepackage{hyperref}
\usepackage{fancyhdr}
\usepackage{setspace}
\usepackage[margin=0.8in]{geometry}
\usepackage{graphicx}
\usepackage{placeins}
\usepackage{textcomp}
\usepackage{tabularx}
\usepackage[normalem]{ulem}
\useunder{\uline}{\ul}{}
\usepackage{acro}
\usepackage{wrapfig}
\usepackage{etoolbox}
\usepackage{caption}
\usepackage{tablefootnote}

\newsavebox{\threesubbox}
\usepackage{dblfloatfix}
\usepackage{stackengine}
\usepackage{amsmath}
\usepackage{multirow}
\usepackage[inline]{enumitem}
\newlist{commalist}{itemize*}{3}
\usepackage{float} % The compiler isn't recognizing the float option 'H'
\setlist[commalist]{itemjoin={{, }}, afterlabel= \unskip{{ }}}
\usepackage{nicematrix}

\usepackage{authblk}

\usepackage{acro}

\DeclareAcronym{NCCT}{
  short=NCCT,
  long=non-contrast Computed Tomography,
}

\DeclareAcronym{nnUNet}{
  short=nnUNet,
  long=No New U-Net,
}

\DeclareAcronym{CNN}{
  short=CNN,
  long= Convolutional Neural Network,
}

\DeclareAcronym{ASPECTS}{
  short= ASPECTS,
  long= Alberta Stroke Program Early CT Score,
}

\DeclareAcronym{AVD}{
  short= AVD,
  long= Absolute Volume Difference,
}
\DeclareAcronym{VS}{
  short= VS,
  long= Volumetric Similarity,
}
\DeclareAcronym{SDT}{
  short= SDT,
  long= Surface Dice at Tolerance,
}
\DeclareAcronym{HD 95}{
  short= HD 95,
  long= Hausdorff Distance 95 percentile,
}

\DeclareAcronym{ASD}{
  short= ASD,
  long= Average Surface Distance,
}

\newcommand{\mycomment}[1]{} % for long comments

\newcommand\Tstrut{\rule{0pt}{2.6ex}}         % = `top' strut
\newcommand\Bstrut{\rule[-0.9ex]{0pt}{0pt}}   % = `bottom' strut
\AtBeginEnvironment{tabular}{\footnotesize}

\newcommand{\beginsupplement}{
  \setcounter{table}{0}  
  \renewcommand{\thetable}{S\arabic{table}} 
  \setcounter{figure}{0} 
  \renewcommand{\thefigure}{S\arabic{figure}}
}

\begin{document}
\raggedright

\title{\textbf{Random Expert Sampling for Deep Learning Segmentation of Acute Ischemic Stroke on Non-contrast CT}}%

\author[1]{Sophie Ostmeier, MD}
\author[1]{Brian Axelrod, PhD}
\author[1]{Benjamin Pulli, MD}
\author[2]{Benjamin F.J. Verhaaren, MD, PhD}
\author[1]{Abdelkader Mahammedi, MD}
\author[1]{Yongkai Liu, PhD}
\author[3]{Christian Federau, MD, MS}
\author[1]{Greg Zaharchuk, MD, PhD}
\author[1,*]{Jeremy J. Heit, MD, PhD}

\affil[*]{corresponding author's email: jheit@stanford.edu}
\affil[1]{Stanford School of Medicine, 453 Quarry Rd, Palo Alto, CA 94304, USA}
\affil[2]{KU Leuven, Medical Imaging Research Center, Leuven, Belgium}
\affil[3]{AI Medical AG, Zurich, Switzerland}

\date{}

\maketitle

\doublespacing

\section*{Short Title}
Random Expert Sampling in Deep Learning for Stroke

\section*{Word Count}
3273 words in total

% Jeremys comments:
% #1
% I think we can clarify in the methods why STAPLE is not appropriate in a sentence, which shouldn’t add too much length. 

% We can also add a sentence about the decision to use 3 raters in the methods. We need 3 as a “tie-breaker” between two raters, and a larger number is not feasible.

% Regarding the other dataset, I think we can just add a sentence about these patients without stroke, and that these patients were included to optimize the model. I’m guessing these were patients with headaches or other non-stroke like symptoms, but I suspect Soren could clarify for us. We can add in the methods that since these were known to not be stroke patients, no segmentation was performed by the expert raters.

% #2

% Regarding DWI images, obviously these would be great, but VERY few people have these data. We could add a sentence to the methods that since the CT data were collected from multiple sites, all patients had NCCT, CTA, and CTP, but not DWI.

% I’ll have to look at the Theranostics paper. If appropriate, we can reference it and discuss it in the discussion is warranted.

\thispagestyle{empty}
\newpage
\setcounter{page}{1}

\section*{Abstract}
\subsection*{Purpose}
To automatically identify the presence and location of ischemic brain tissue on Non-Contrast CT 

\subsection*{Materials and Methods}

The data set consisted of 260 Non-Contrast CTs from 233 patients of acute ischemic stroke patients recruited in the DEFUSE 3 trial.  
A benchmark U-Net was trained on the reference annotations of three experienced neuroradiologists to segment ischemic brain tissue using majority vote and random expert sampling training schemes. 
We used a one-sided Wilcoxon signed-rank test on a set of segmentation metrics to compare bootstrapped point estimates of the training schemes with the inter-expert agreement and ratio of variance for consistency analysis. 
We further compare volumes with the 24h-follow-up DWI (final infarct core) in the patient subgroup with full reperfusion and we test volumes for correlation to the clinical outcome (mRS after 30 and 90 days) with the Spearman method.

\subsection*{Results}
Random expert sampling leads to a model that shows better agreement with experts than experts agree among themselves and better agreement than the agreement between experts and a majority-vote model performance (Surface Dice at Tolerance 5mm improvement of 61\% to 0.70±0.03 and Dice improvement of 25\% to 0.50±0.04). The model-based predicted volume similarly estimated the final infarct volume and correlated better to the clinical outcome than CT perfusion.

\subsection*{Conclusion}
A model trained on random expert sampling can identify the presence and location of acute ischemic brain tissue on Non-Contrast CT similar to CT perfusion and with better consistency than experts. This may further secure the selection of patients eligible for endovascular treatment in less specialized hospitals.
%\linenumbers
\pagebreak[4]
\section{Introduction }
Stroke is a leading cause of death and disability globally \cite{feigin2021global, LakomkinPrevalence}. Timely treatments, such as thrombolysis or thrombectomy, represent effective treatment options to improve clinical outcomes for patients suffering from acute ischemic stroke \cite{HuoLargeCore, SarrajLargeCore, Nogueira6to24, hacke2008thrombolysis}. 
Non-Contrast CT (NCCT) is widely used to differentiate ischemic from hemorrhagic stroke.
It can be further used to estimate the extent of irreversibly damaged brain tissue as an imaging biomarker for the appropriate triage of patients \cite{KimUtilization, McDonoughState}.

Advances have been made to facilitate the delineation of ischemic brain tissue with ruled-based algorithms for perfusion imaging \cite{Albers6to16}. But resources, specialized technicians, and physicians are needed for interpretations.
%and spatial agreement compared to the final ischemic core is poor \cite{nguyen2022noncontrast,sarraj2022accuracy}.
The most utilized (96-100\%) and cheapest imaging modality for acute stroke patients is the NCCT \cite{KimUtilization, WangSocioeconomic}.

The ASPECT score is an established semi-quantitative method to evaluate treatment eligibility in ischemic stroke patients based on NCCT by dividing the affected hemisphere into 10 structural regions \cite{ HuoLargeCore, SarrajLargeCore}. 
However, the extent and relative quantity of damaged brain tissue within a region to be rendered has not been defined \cite{phan2006aspects, SchroederCritical}.
The individual regions cover different amounts of brain tissue which results in their unequal weighing into the score.  
The location of the stroke has a minor role in the ASPECT score value. Although it has been shown that the location of the stroke estimates the long-term effect on patients' outcome more precisely \cite{weaver2021strategic, wang2019stroke, meyer2016voxel, broocks2018computed}. 

A segmentation tool for acute ischemic brain tissue on NCCT may not only accurately quantify the volume but also allows the mapping of the stroke location to symptoms to better guide treatment decisions. 
However, a low signal-to-noise ratio limits rule-based algorithms and expert neuroradiologists to segment ischemic stroke on NCCT. An automated pipeline is therefore of high interest \cite{nowinski2020ischemic,chen2022prognosis,el2022evaluating}.

Supervised deep learning models have been widely applied for the segmentation of medical images and show promise for applications in stroke imaging, but rely on accurate expert reference annotations segmentation during training \cite{LiuDetection,yu2020use}. 
To enhance the accuracy of ground truth segmentation, multiple expert annotations are commonly fused with a majority or with the probabilistic fusion algorithm STAPLE. The result is a binary ground truth (a voxel is part of the ischemic core or not). Note that STAPLE is not applicable in the context of this study, where sometimes only two, one, or no experts agree on ischemic stroke segmentation \cite{warfield2004simultaneous}.
Instead, we use random expert sampling, meaning randomly choosing an expert segmentation as ground truth for each training example. We then compare the model's performance to the standard majority vote. 
We hypothesize that randomly sampling experts during training is better in approximating the ground truth of the ischemic core better than a majority vote due to its ability to encode the true probability when using only binary expert labels. 
For ischemic brain segmentation, the true probability of neurons being irreversibly damaged by ischemia maps to a continuous function of time, collateral blood flow, and clinical variables \cite{saver2006time, marks2014effect, johnston2000predictive}. 
If we consider a Bernoulli random variable corresponding to a single pixel (a coin flip with probability $p$) is defined as 1 with probability $p$ and 0 with probability $1-p$. 
Consider this variable with $p=50\%$ with three samples corresponding to the annotations of three experts that happen to be $(0,0,1)$. 
If we used a majority vote pre-processing scheme, this would be summarized as a single zero, and any resulting model or estimator would attempt to predict a zero. 
However, maximizing the cross entropy loss function of the deep learning model (identical to taking the maximum likelihood estimate, (Supplemental Material, Section \ref{sec:cross_entropy}) would result in an estimate of $1/3$. In fact, we'd expect a better estimate in 75\% of three-expert datasets. 
Finally, we correlate the prediction of the random expert model to the clinical outcome and compare it to CT perfusion and follow-up MR imaging.

\section{Methods}

\subsection{Study Design and Data}
In this post hoc analysis of the randomized DEFUSE 3 trial, 260 Head NCCT examinations from 233 patients with enrollment between May 2016 through May 2017 were included. 
The primary trial outcome determined thrombectomy eligibility for patients with acute ischemic stroke with an onset time within 16 hours. Detailed information of the inclusion and exclusion criteria can be found in \url{https://clinicaltrials.gov/ct2/show/NCT02586415} and \cite{Albers6to16}. 
Furthermore, 156 consecutive patients without an ischemic stroke from a different institutional database were included to evaluate the image-classification task (relevant stroke vs. no relevant stroke volume). These patients were screened  for  stroke  (from  2011  to  2017) and confirmed to have no stroke with a normal DWI scan 24h after the initial NCCT, but were diagnosed with an alternative condition on the basis  of  all  available  clinical  and  radiologic  data  available  to  the treating physician \cite{ChristensenOptimizing}. 
We included patients from the University Hospital Lausanne (UHL) (1723 patients) as an external generalization cohort. We randomly chose 35 patients of this cohort where each patient was admitted between 2016 through 2019, had an onset time between 3h to 24h and an initial NCCT study.
Institutional review board approval was obtained, and consent was waived.

\subsection{Image Reference Annotations}
Three experienced fellowship-trained neuroradiologists (experts A, B, and C) with 4, 4, and 5 years of post-fellowship experience (A.M., B.F.J.V., J.J.H.). 
Manual segmentations were done in Horos (Horosproject.org, version 4.0.0). 
The experts were given the information that each patient had a large vessel occlusion, but the side of the vessel occlusion was not provided. 
If no abnormal hypodensity was found, experts were offered the option of no segmentation. All manual segmentations were checked for correct coregistration to the NCCT. 
For the data set of the healthy patient, an empty segmentation mask was generated.
The external generalization cohort was segmented by one fellowship-trained neuroradiologists with 2 years of post-fellowship training. 

\subsection{Data Preparation for Training}
\begin{figure}[h]
    \centering
    \includegraphics{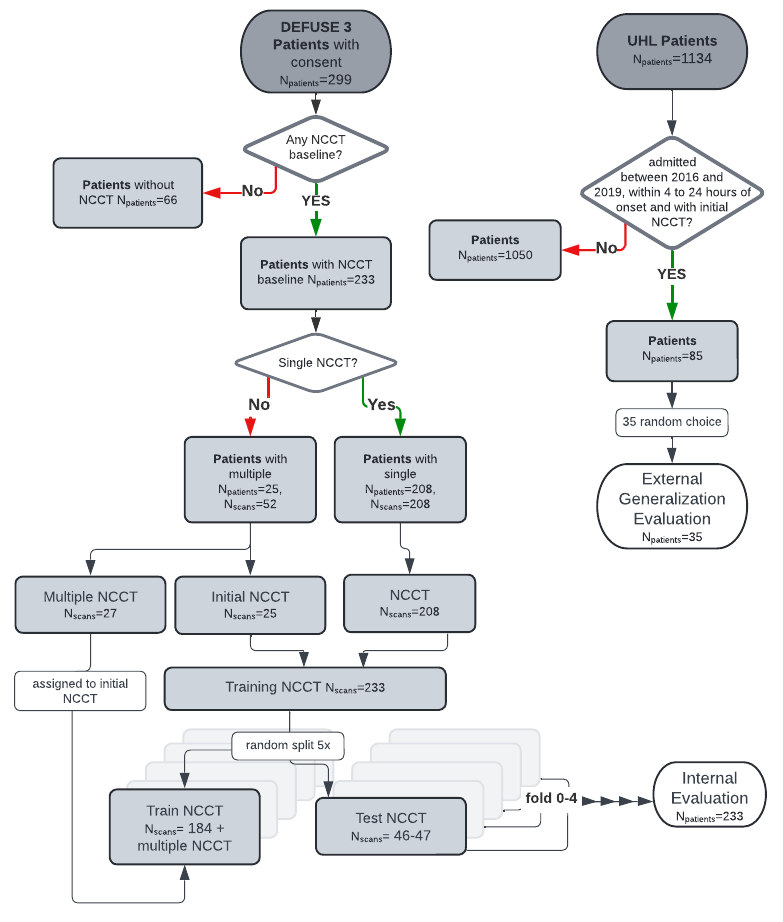}
    \caption{Flowchart of the data partition. 233 patients with their initial NCCT were randomly split into 5 folds of training and test sets. 25 patients had multiple NCCT. Those were only assign to the initial NCCT when in the training set. The external generalization cohort included 35 patients.}
    \label{fig:flowchart}
\end{figure}

The cohort of 233 patients was randomly split into five subgroups for five-fold cross-validation with 208 patients for the training, and 52 for the test sets. If a patient had undergone multiple baseline NCCT (n=27 patients), the additional scans were only added to the corresponding patient NCCT in the training set to prevent information leakage (Figure \ref{fig:flowchart}). The 156 healthy cases were randomly and equally added to the test sets (n=33 per test set) with empty ground truth annotations.

\subsection{Model Configuration and Training \label{sec:Training}}
\begin{figure}[h]
    \centering
    \includegraphics[width=\textwidth]{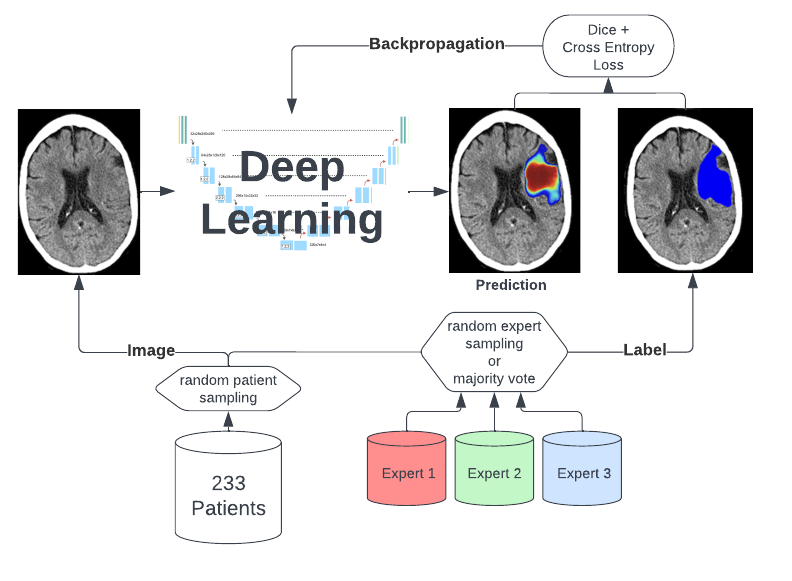}
    \caption{Training scheme pipeline with sampling strategy for random expert sampling and majority vote}
    \label{fig:training}
\end{figure}

A preprocessing pipeline and data loader were created to allow on the fly training with majority vote and random expert sampling (Figure \ref{fig:training}).
A majority vote means to sum the reference masks of all experts, set the voxels with a value greater than half the number of experts to one, and all other voxels to zero.
Random expert sampling means choosing an expert for each training example with an equal probability.

\begin{figure}
    \centering
    \includegraphics{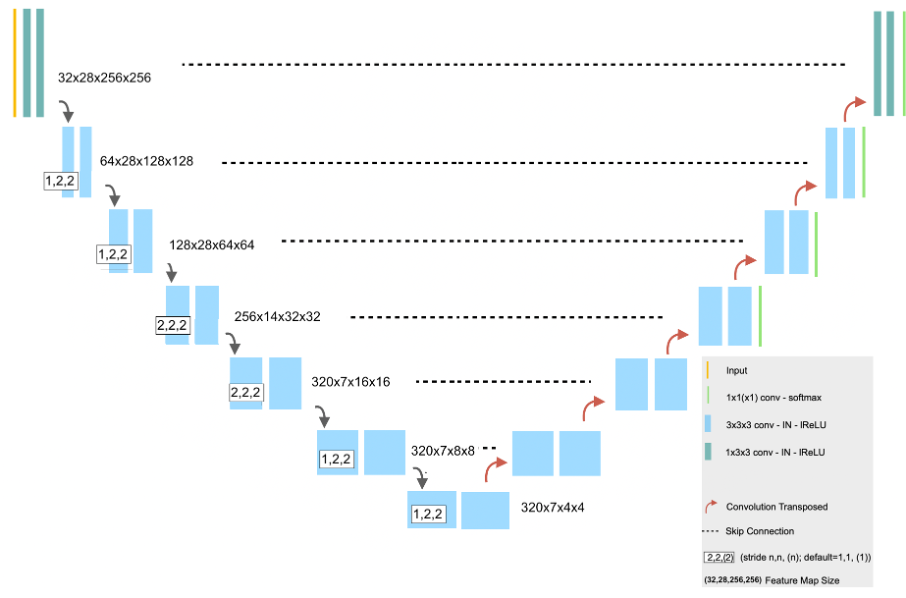}
    \caption{\bfseries  Model Architecture}
    \label{fig:archi}
\end{figure}

A baseline nnUNet architecture and default hyperparameter include a patch size of 28x256x256 and spacing of (3.00, 0.45, 0.45), 7 stages with two 3D convolutions per stage, leaky ReLU as activation function, Dice + Cross-Entropy loss function, a batch size of 2, SGD optimizer with 0.99 Nestov momentum and He initialization (Section \ref{sec:supp_mat}, Fig. \ref{fig:archi}). 
The model was trained with NCCT as input and manual annotations of experts as ground truth to output a predicted segmentation mask of acute ischemic stroke on NCCT \cite{IsenseennU-Net}. 
As all models share the same core nnUNet component and for fairness and ease of comparability, we let all models undergo the same training schedule with default hyperparameters to prevent information leakage.

\subsection{Model Testing}
All analyses have been performed on the aggregated test sets of the five folds. This procedure is explored and validated in previous literature \cite{cawley2010over}.  The results of the aggregated test sets (n=389) were tested for normal distribution with the Shapiro and shown as median and 95\% CI estimated by the bootstrapping (R=1000)\cite{lakens2017equivalence}
We tested the best-performing model (overall highest metrics value) on the external generalization cohort.
 
\subsection{Outcome Measures and Statistical Analysis}
%We developed a score to quantify the uncertainty of the reference annotations to offer an indicator for future data set, when random expert sampling should be chosen over the majority vote. Therefore, we computed the pixel-wise entropy across experts. 

R (Version 2022.02.3) was used for statistical analysis.

\begin{figure}[h]
    \centering
    \includegraphics[width=\textwidth]{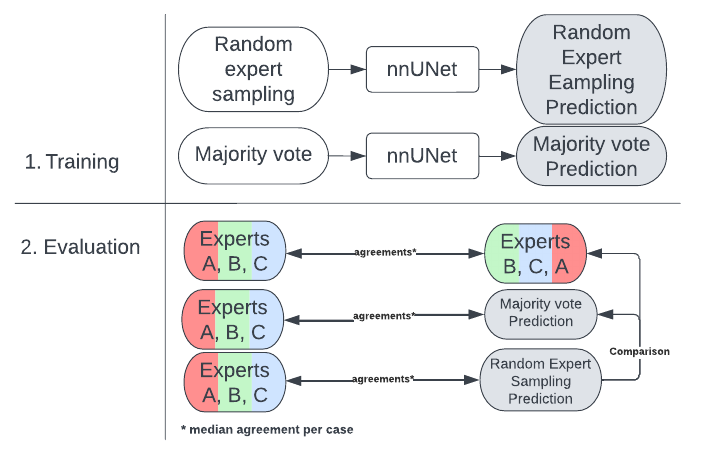}
    \caption{First two model were trained on majority vote and random expert sampling. Second, the median agreement per case for inter-expert agreement, model-expert agreement for the prediction of majority vote and random expert sampling was the basis to compare random expert sampling to the majority vote and inter-expert agreement. }
    \label{fig:analysis}
\end{figure}

All three expert reference annotations per NCCT were compared to the inter-expert agreement, to the prediction of the majority vote model, and to the prediction of the random expert sampling model (Figure \ref{fig:analysis}). The per-patient median was used for further analysis.

The model-based predictions were evaluated for segmentation error, image classification, volume classification, correlation to CT ischemic core, 24h-follow-up volume, and clinical outcome prediction.

To access the segmentation and image-classification task a clinically motivated threshold of 1ml was chosen to differentiate between relevant stroke and no relevant stroke volume. 
For example, after determining no relevant stroke volume on the NCCT, the location, and volume of ischemic brain tissue of 0.5 ml are unlikely to influence the triage of patients and excluded from the segmentation evaluation. Expert A, B, and C had 155, 226, and 206 manual segmentations $>1$ml and 78, 7, and 27 manual segmentations $<1$ml, respectively.

All NCCTs with a median reference annotation above 1ml were evaluated for segmentation error with the following metrics. 
\begin{itemize}
\item Volume-based metrics (Volumetric similarity and absolute volume difference [ml])
\item Overlap metrics (Dice, Precision and Recall)
\item Distance metrics Hausdorff distance with the 95 percentile [mm], and the surface dice at tolerance 5mm
\end{itemize}
The definition can be found in the supplemental material \ref{sec:supp_mat}, Table \ref{table:metric_definition} \cite{ostmeier2022evaluation}.

The Surface Dice at Tolerance measures the distances between each surface voxel of the reference and predicted mask. The chosen tolerance allows a maximum distance between surface voxels on the reference and predicted masks to be considered true positive voxels. We use this metric to account for possible pathophysiological and modality-related lower signals that cause more variability in the outer compared to the inner region of the ischemic stroke \cite{nikolov2018deep, Maier_Hein_pitfall_metric}.

For the image-classification task, the predictions were categorized in stroke volume above or below 1ml including the healthy cases. 
Sensitivity, specificity, F-score, Correct Classification Ratio, and Area under the curve served as metrics. 

We used each metric to evaluate how close the model-based predictions were to experts (accuracy) and how consistent the model-based predictions were across the patient population (precision).  

Accuracy was tested for statistical superiority with a one-sided Wilcoxon sign rank test and a lower boundary of 0.0 ($p<0.05$, Figure \ref{fig:analysis}).
We evaluate for precision with the ratio of variances and standard error (bootstrapped, R=1000). We then tested for statistical superiority with a delta of 0.0.

To evaluate for ischemic core volume size that alters clinical treatment decisions based on ANGEL-ASPECT and SELECT-2 \cite{SarrajLargeCore,HuoLargeCore}, the predicted segmentations were classified in $<$1ml, $<$50ml, $<$100ml and $>$100ml and compared to the median expert volume with Cohen's and Fleiss' kappa.

To put the results into perspective, we computed the Spearman correlation coefficient of the ASPECT and ischemic core volumes estimated by experts, deep learning models, and CT perfusion to the clinical outcomes.

We compared the ischemic core volumes of the random expert sampling model and CT perfusion (30\% CBF) to the 24h-follow-up DWI (final infarct core) with Bland-Altman plots in patients with full reperfusion (TICI $\geq$2B). We analyzed differences in correlation coefficients with Fisher's z-test.

All p-values were adjusted for multiple comparisons with the Holm-Bonferroni method.

\subsection{Model and Data Set Availability}
The model and data set are available upon request to the corresponding author.

\section{Results}
\subsection{Patient Characteristics}
We analyzed 233 randomized and non-randomized patients in the DEFUSE 3 trial (121 women (52\%), median (IQR) age, 69 (59-78) years). 
The expert ischemic core volume and ischemic core (CBF$<=$30\%) and penumbra (Tmax $.=$6) volume with perfusion imaging were 8[3-26]ml, 11[2-28] and 104[62-157] ml, respectively (median [IQR] ml). 
The median onset to image time of the 50 patients witnessed was 10[8-12]h. Further patient characteristics are summarized in Table \ref{tab:pcharacteristics}.

\subsection{Model Performance}
The model-based segmentation with the random expert sampling showed statistically significant better agreement with experts than 
inter-expert agreement and majority vote training scheme (median $\pm$ 95\% CI (bootstrapped), Surface Dice at Tolerance 0.7$\pm$0.03 vs. 0.56$\pm$0.03, 0.68$\pm$0.05, Dice 0.50$\pm$0.04 vs. 0.31$\pm$0.04, 0.29.$\pm$0.01, Absolute Volume Difference of 5.36$\pm$1.32 ml vs. 10.20$\pm$2.09 ml, 7.91$\pm$1.77 ml) (Table \ref{tab:segmentation}). 

The consistency of the model-based segmentation with the random expert sampling across the patient population was significantly better for VS, AVD, and Precision but insignificant elsewhere. However, most non-significant comparisons have a standard deviation that includes variance ratios of 1. For these metrics, no conclusion about the segmentation precision superiority or inferiority can be made and the variance is most likely similar.

\begin{figure}[h]
    \centering
    \includegraphics[width=\textwidth]{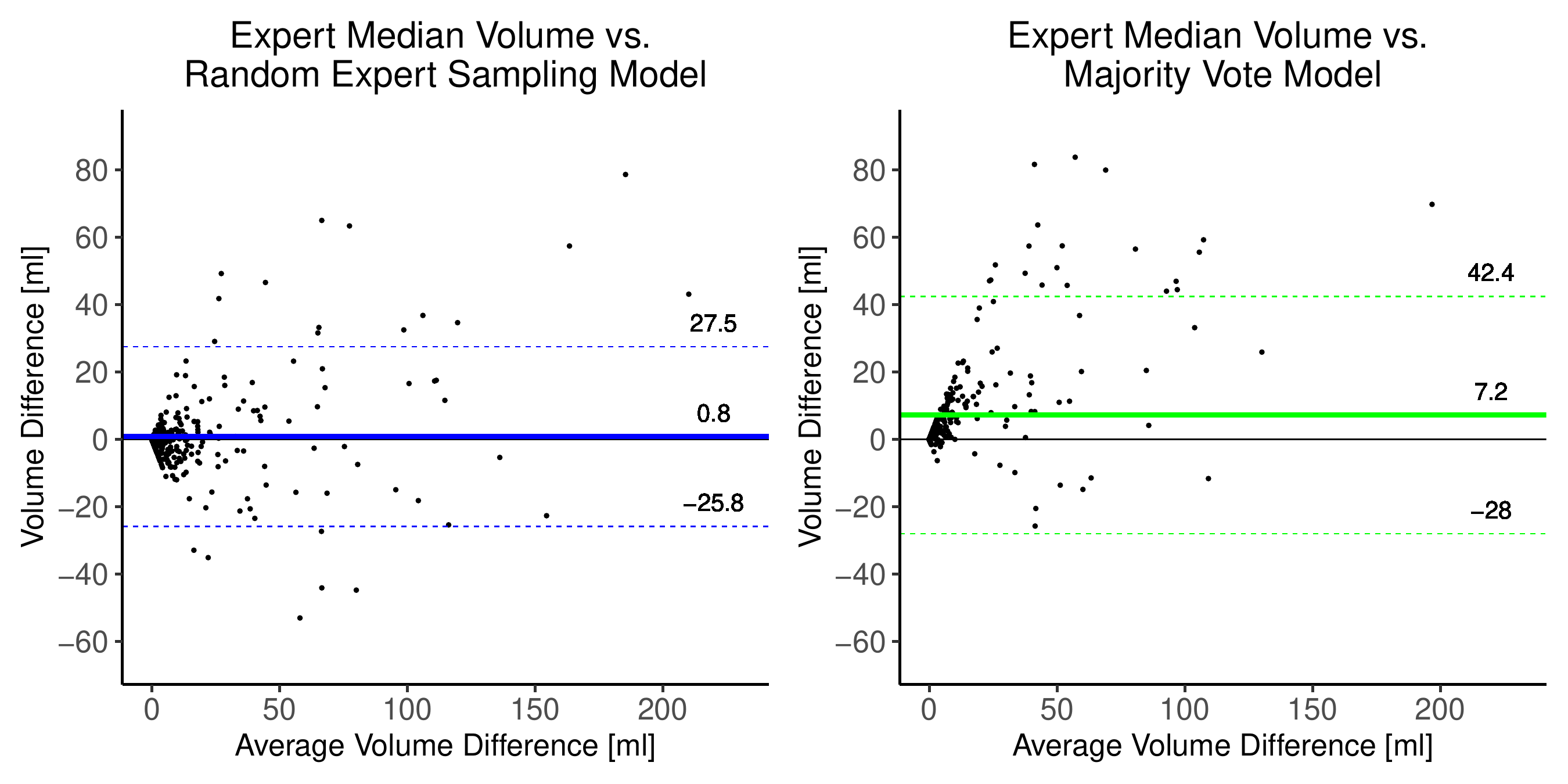}
    \caption{Bland-Altman for Random Expert Sampling (blue) and Majority Vote Model Volume  (green) estimates compare to Median Expert Volume.
    %when compared to the median expert volume and CTP ischemic core volumes.
    }
    \label{fig:blandexpert}
\end{figure}

The model-based segmentation with the random expert sampling showed a smaller average volume difference and 95\% CI when compared to the median expert volume with Bland-Altman plots, while the majority vote tends to underestimate the median expert volume (Figure \ref{fig:blandexpert}) (mean and 95\% CI volume difference of random expert sampling vs. majority vote, 0.8(-25.8-27.5) ml vs. 7.2(-26-42.4) ml).

\begin{figure}[h]
    \centering
    \includegraphics[width=\textwidth]{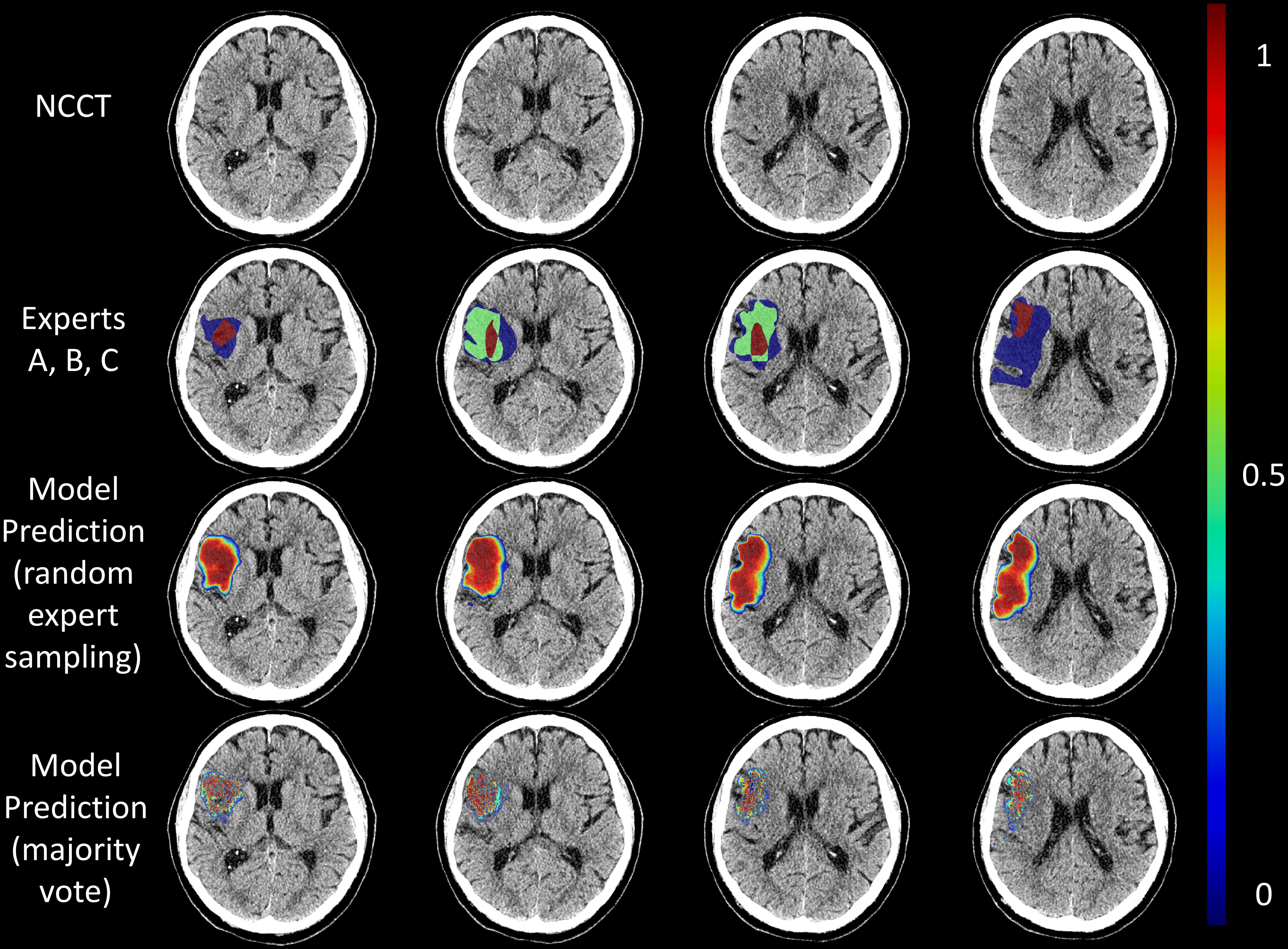}
    \caption{Example of a heatmap of expert compared to models trained on random expert sampling and majority vote. Low values are represented by blue to green colors, and high values by yellow to red colors. The prediction of a majority vote underestimates acute ischemic brain tissue shown.}
    \label{fig:example}
\end{figure}

Qualitatively, the probability predictions of random expert sampling correlate better to the fields segmented by the experts than to the predictions for the majority vote which may show overfitting to the single hypodense voxels and image noise (Figure \ref{fig:example}). 

The random expert sampling model indicated similar performance to classifying cases into relevant and not relevant stroke volume vs. inter-expert agreement and the majority vote model, respectively (AUC 0.92$\pm$0.02 vs.  0.93$\pm$0.02,  0.90$\pm$0.02 ) (Table \ref{tab:classification})

\subsection{Evaluation of NCCT ischemic Core Volume as Imaging Biomarker in Clinical Practice}
Cohen's kappa for volume classification ($<$1ml, $<$50ml, $<$100ml and $<$100ml) of the random expert sampling model is higher than Fleiss' kappa for inter-expert agreement and Cohen's kappa for the majority vote model (0.52 $\pm$0.06 vs. 0.32 $\pm$0.04 and 0.24 $\pm$0.05).

When comparing the initial ischemic core volume to the clinical outcome (mRS after 30 and 90 days), random experts sampling model volume and ASPECTS correlate significantly to the mRS scores after 30 and 90 days and the CTP volume does not (Table \ref{tab:clinical_outcome}).

\begin{figure}[h]
    \centering
    \includegraphics[width=\textwidth]{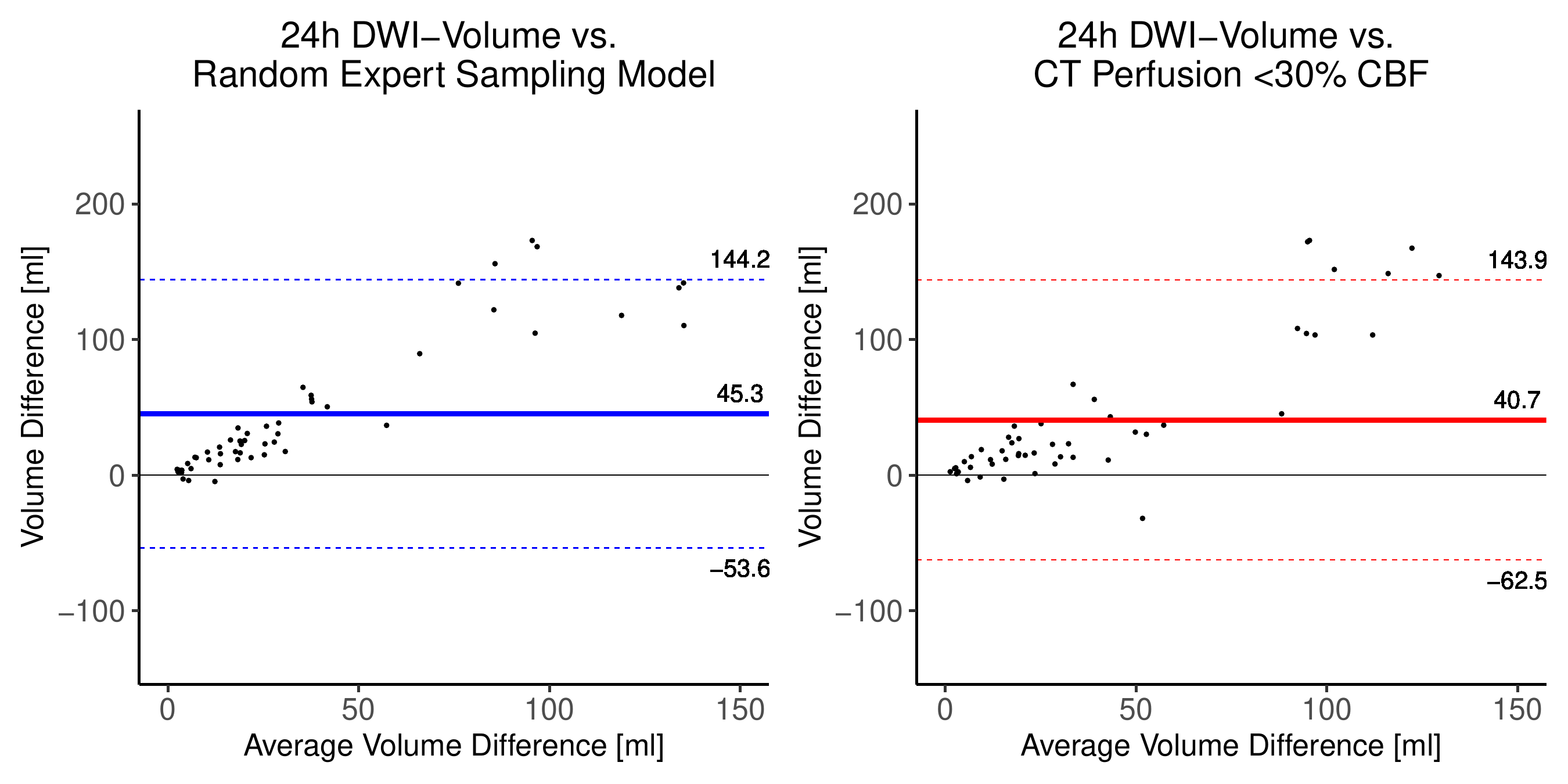}
    \caption{Bland-Altman for Random Expert Sampling Model Volume (blue) and CTP Ischemic Core Volume <30\% (red) compared to 24h DWI-Volume for all reperfusors (TICI$\geq$2B).
    }
    \label{fig:blandexpertDWI}
\end{figure}

\begin{figure}[h]
    \centering
    \includegraphics[width=\textwidth]{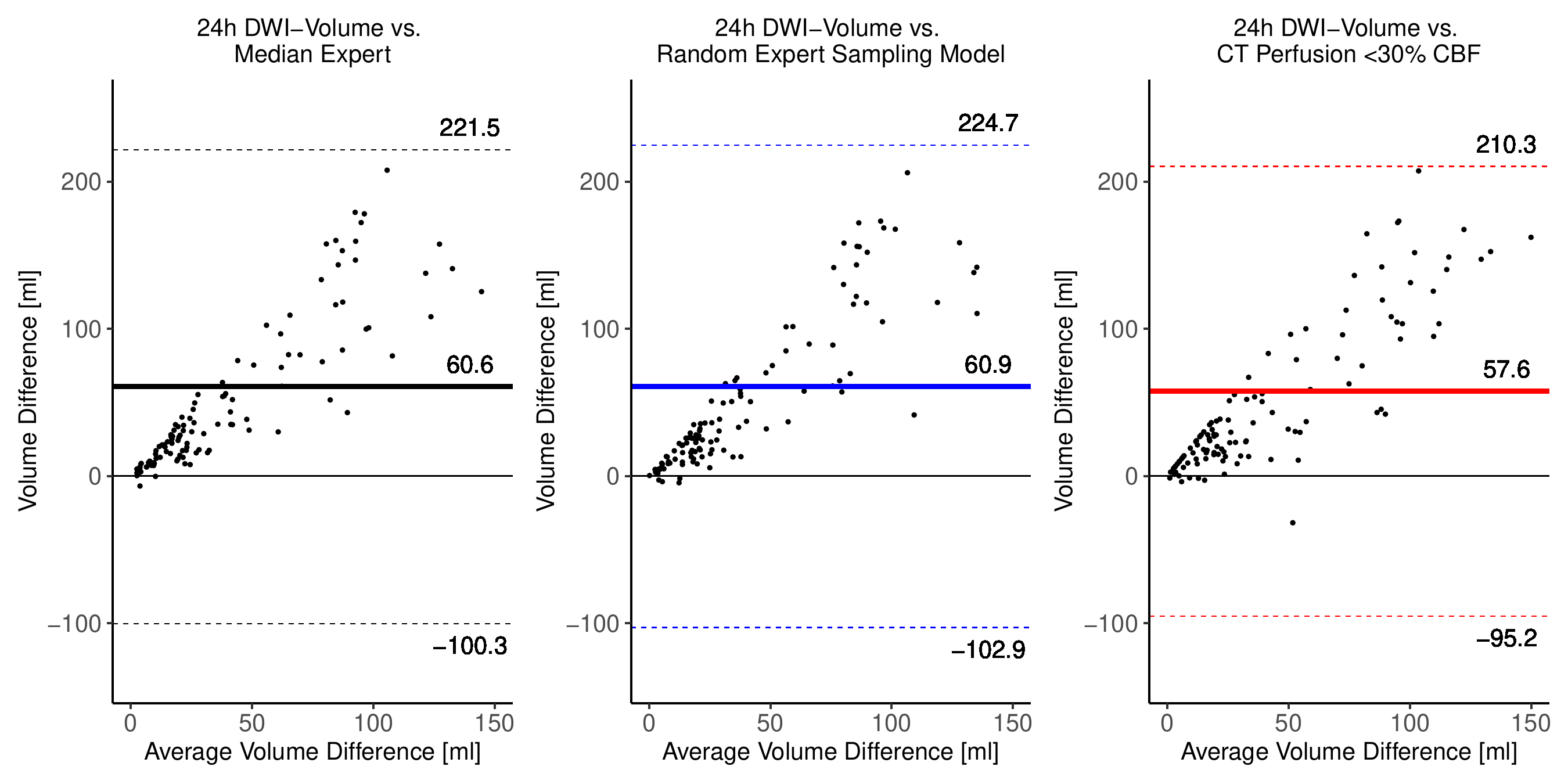}
    \caption{Bland-Altman for Median Expert Volume (black), Random Expert Sampling Model Volume (blue) and CTP Ischemic Core Volume $<$30\% (red) compared to 24h DWI-Volume for over entire patient population).
    }
    \label{fig:blandexpertDWI_all}
\end{figure}

Subgroup analysis of patients with full reperfusion (TICI $\geq$2B, n=51) shows no significant difference when comparing the average volume differences for the random expert sampling volume prediction and CT perfusion volume to the 24h-follow-up DWI volume (Fisher's z-test p-value = 0.355, Figure \ref{fig:blandexpertDWI}). Analysis of the whole patient population shows similar results (Supplemental Material, Figure \ref{fig:blandexpertDWI_all}).

\section{Discussion} 
In this study, we found that random expert sampling training of a benchmark deep learning model leads to better agreement with experts than experts among themselves for the segmentation of the ischemic core on NCCT. 
The model-based ischemic core shows similar volume agreement with the final infarct volume and better correlation to the clinical outcomes as CT perfusion.

The significance of these findings is threefold. 
First, the model outperforms experts of the highest human expertise when jointly trained by the experts themselves. 
Second, the model's volume demonstrates similar clinical predictive value as the ischemic core volume estimates of current rule-based algorithms.
Third,  we found that contrary to prior deep learning applications, fusion techniques such as a majority vote cause the model to overfit to single voxel values and segment less meaningful clinical information \cite{vincent2021impact, Langerak2008LabelFusion,Landman2012StatisticalFusion, warfield2004simultaneous}.

ECASS I, ESCAPE, DAWN, DEFUSE 3, SELECT-2 and ANGEL-ASPECT are clinical trials that have included the measures of ischemic core on NCCT as inclusion and treatment criteria for patients with acute ischemic stroke \cite{hacke2008thrombolysis, goyal2015randomized, Albers6to16,Nogueira6to24, SarrajLargeCore,HuoLargeCore}.

The ECASS I and TWIST trial specified hypodensity consistent with ischemic brain tissue of more than 1/3 of the medial cerebral artery territory as exclusion criteria\cite{hacke1995intravenous,zha2022non}.
Equivalently, ESCAPE and (IMS)-III Trial and retrospective studies suggest \ac{ASPECTS} of  $<5$, $<6$, $<7$ or $<8$ for estimating a large ischemic infarct and to guide endovascular treatment decisions \cite{goyal2015randomized, hill2014alberta, zha2022non, yoo2014impact}.
However, the benefits of thrombectomy have recently also been demonstrated for patients with ASPECTS $3-5$ in SELECT-2 and ANGEL-ASPECT \cite{HuoLargeCore,SarrajLargeCore}.
Although \ac{ASPECTS} is widely used as a predictor for patients to benefit from endovascular treatment using different thresholds, it is limited by inter-rater variability, a modest correlation to ischemic core volumes and location \cite{barber2000validity,GoyalEndovascular,BarberValidity,SchroederCritical}.

Besides the ASPECT score, the ASPECT-ANGEL and SELECT-2 trials have also included patients with an ischemic core volume of $70-100$ml and $>5$0ml that was quantified with CT perfusion \cite{HuoLargeCore,SarrajLargeCore}. 

We proposed a deep learning segmentation method of the ischemic core on NCCT that may alleviate limitations of perfusion imaging and the \ac{ASPECTS} score.

Different reference standard definitions of the ischemic core exist across various stroke imaging modalities (MR and CT Perfusion, DWI).
Sarraj et. al and SELECT investigators have shown that the ischemic core volume on CT perfusion is biased and poorly overlaps with the final infarct volume (median (IOR) absolute volume difference 17.9 (5.6–43.9)ml and Dice score 0.1 (0.0-0.4)) \cite{sarraj2022accuracy}.

Diffusion-weighted imaging (DWI) within a short time after the NCCT is used in prior works as ground truth for the segmentation or detection of the ischemic core on NCCT \cite{KuangSegASPECT,QiuNCCTMachine,gauriau2023head, lu2022identification}. 
However, especially in earlier time windows ($<1$h) cytotoxic edema is the predominant image abnormality depicted as diffusion restriction on DWI \cite{albers1998diffusion}. 
Vasogenic edema depicted as hypodensity on \ac{NCCT} develops $>1-4$h after stroke onset and suggests irreversibly damaged brain tissue (ischemic core) ~\cite{almandoz2011imaging}. 
The image correlation of the underlying ischemic core pathology of a DWI lesion on NCCT especially in the earlier time frames seem unclear and voxel-wise comparisons of NCCT and DWI has been scarcely studied \cite{lansberg2000comparison}. 

%Prior deep learning applications focus on returning calibrated uncertainty estimates to inform clinicians about the model confidence of its prediction, and model uncertainty in \cite{czolbe2021segmentation,lemay2022label}. 
%Kohl et al. propose the probabilistic U-Net to encode inter-rater uncertainty and to provide clinicians distributions over possible segmentations rather than point estimates \cite{kohl2018probabilistic}. 
%Acute stroke imaging is time-sensitive. A model that directly predicts a probability heatmap or a binary segmentation seems more useful than proposing multiple possible segmentations which need time and expertise to evaluate. 

Deep learning medical image segmentation is more often supervised by human experts' annotations and fusion methods (e.g. majority vote) may approximate an error-free ground truth when collecting annotations from multiple experts. 
In a segmentation task with reference annotation of uncertainty, small target lesions, or empty segmentations reference annotations require at least highly skilled experts to minimize the errors, and multiple experts' segmentations are needed to approximate the distribution of interpretations, which is resource-intensive and timely \cite{ostmeier2022evaluation}.
For acute ischemic stroke segmentation on NCCT, fusion methods for categorical voxel classes may limit the ability of the model to learn segmentation tasks where experts inherently disagree. We found that advanced fusion methods, such as STAPLE or SIMPLE are not applicable and tested majority voting with modest performance \cite{warfield2004simultaneous,Langerak2008LabelFusion}. 

We propose a training requiem for multi-expert training that maximizes the encoded information of the NCCT interpretations from three expert neuroradiologists.  We used three experts so that there is a “tie-breaker” between two raters. A larger number of experts is not feasible. The model trained with this combined encoded information (random expert sampling methods) agreed more with experts when compared to the inter-human-reader agreement (Dice score 0.51±0.04 vs. 0.31±0.04 (Table \ref{tab:segmentation})). When comparing the model-based volume and CT Perfusion volume to the final infarct volume we could find no significant difference in final infarct estimation.

Neuroimaging is a significant cost driver in acute ischemic stroke patients \cite{christensenCostDriver}. 
In addition, reduced imaging modalities and scan time may increase the net monetary benefit of earlier endovascular treatment (\$10,593 for every 10 minutes) \cite{Straka_time,kunz_monetary_benefit_2020}. However, additional and prospective studies regarding clinical outcomes are needed before the long-term cost-effectiveness of using NCCT and deep learning tools can be concluded \cite{nguyen_non_2022}.

This study has limitations. 
First, we only use a patient cohort of one multicenter clinical trial with the primary outcome measure of selecting patients that present within 16h. For the image classification, we do not include stroke mimics or hemorrhagic stroke patients. 
Further research may test the proposed method in a broader and prospective patient population. 
Second, we include three experts. The distribution of image interpretation may require a larger set of neuroradiologists which would be resource-intensive but may improve performance.
Third, the volume of the stroke is only one imaging feature that correlates with the clinical outcome and may influence decision-making. In future work, deep learning methods may offer voxel-based lesion-symptom mapping for better estimation of clinical outcome prediction.

In clinical practice, our model may identify, quantify, and localize acute ischemic stroke on NCCT comparable to CT perfusion. An accurate and precise deep learning segmentation methodology allows less specialized on-call clinicians to utilize the information for treatment decisions given by \ac{NCCT}.
This possibly enhances the impact of NCCT in endovascular treatment decisions for ischemic strokes as a basic, cheap, and widely available imaging modality. 

\section{Acknowledgments}
-

\section{Sources of Funding}
Stanford School of Medicine

\section{Disclosures}
Sophie Ostmeier: none\\
Brian Axelrod: none\\
Benjamin F.J. Verhaaren: none\\
Abdelkader Mahammedi: none\\
Benjamin Pulli: none \\
Yongkai Liu: none\\
Christian Federau: Founder and CEO of AI Medical AG\\
Greg Zaharchuk: co-founder, equity of Subtle Medical, funding support GE Healthcare, consultant Biogen
Jeremy J. Heit: Consultant for Medtronic and MicroVention, Member of the medical and scientific advisory board for iSchemaView\\

\section{Supplemental Material}
Supplemental Methods \\
Tables S1–S2\\
Figure S1\\

\pagebreak[4]
\section{References}
\bibliographystyle{IEEEtran}
\bibliography{refs_here}

\begin{thebibliography}{10}
\providecommand{\url}[1]{#1}
\csname url@samestyle\endcsname
\providecommand{\newblock}{\relax}
\providecommand{\bibinfo}[2]{#2}
\providecommand{\BIBentrySTDinterwordspacing}{\spaceskip=0pt\relax}
\providecommand{\BIBentryALTinterwordstretchfactor}{4}
\providecommand{\BIBentryALTinterwordspacing}{\spaceskip=\fontdimen2\font plus
\BIBentryALTinterwordstretchfactor\fontdimen3\font minus
  \fontdimen4\font\relax}
\providecommand{\BIBforeignlanguage}[2]{{%
\expandafter\ifx\csname l@#1\endcsname\relax
\typeout{** WARNING: IEEEtran.bst: No hyphenation pattern has been}%
\typeout{** loaded for the language `#1'. Using the pattern for}%
\typeout{** the default language instead.}%
\else
\language=\csname l@#1\endcsname
\fi
#2}}
\providecommand{\BIBdecl}{\relax}
\BIBdecl

\bibitem{feigin2021global}
V.~L. Feigin, B.~A. Stark, C.~O. Johnson, G.~A. Roth, C.~Bisignano, G.~G.
  Abady, M.~Abbasifard, M.~Abbasi-Kangevari, F.~Abd-Allah, V.~Abedi
  \emph{et~al.}, ``Global, regional, and national burden of stroke and its risk
  factors, 1990--2019: a systematic analysis for the global burden of disease
  study 2019,'' \emph{The Lancet Neurology}, vol.~20, no.~10, pp. 795--820,
  2021.

\bibitem{LakomkinPrevalence}
\BIBentryALTinterwordspacing
N.~Lakomkin, M.~Dhamoon, K.~Carroll, I.~P. Singh, S.~Tuhrim, J.~Lee, J.~T.
  Fifi, and J.~Mocco, ``Prevalence of large vessel occlusion in patients
  presenting with acute ischemic stroke: a 10-year systematic review of the
  literature,'' \emph{Journal of NeuroInterventional Surgery}, vol.~11, no.~3,
  pp. 241--245, 2019. [Online]. Available:
  \url{https://jnis.bmj.com/content/neurintsurg/11/3/241.full.pdf}
\BIBentrySTDinterwordspacing

\bibitem{HuoLargeCore}
X.~Huo, G.~Ma, X.~Tong, X.~Zhang, Y.~Pan, T.~N. Nguyen, G.~Yuan, H.~Han,
  W.~Chen, M.~Wei, J.~Zhang, Z.~Zhou, X.~Yao, G.~Wang, W.~Song, X.~Cai, G.~Nan,
  D.~Li, A.~Y.-C. Wang, W.~Ling, C.~Cai, C.~Wen, E.~Wang, L.~Zhang, C.~Jiang,
  Y.~Liu, G.~Liao, X.~Chen, T.~Li, S.~Liu, J.~Li, F.~Gao, N.~Ma, D.~Mo,
  L.~Song, X.~Sun, X.~Li, Y.~Deng, G.~Luo, M.~Lv, H.~He, A.~Liu, J.~Zhang,
  S.~Mu, L.~Liu, J.~Jing, X.~Nie, Z.~Ding, W.~Du, X.~Zhao, P.~Yang, L.~Liu,
  Y.~Wang, D.~S. Liebeskind, V.~M. Pereira, Z.~Ren, Y.~Wang, and Z.~Miao,
  ``Trial of endovascular therapy for acute ischemic stroke with large
  infarct,'' \emph{New England Journal of Medicine}, p. null, 2023.

\bibitem{SarrajLargeCore}
\BIBentryALTinterwordspacing
A.~Sarraj, A.~E. Hassan, M.~G. Abraham, S.~Ortega-Gutierrez, S.~E. Kasner,
  M.~S. Hussain, M.~Chen, S.~Blackburn, C.~W. Sitton, L.~Churilov,
  S.~Sundararajan, Y.~C. Hu, N.~A. Herial, P.~Jabbour, D.~Gibson, A.~N.
  Wallace, J.~F. Arenillas, J.~P. Tsai, R.~F. Budzik, W.~J. Hicks, O.~Kozak,
  B.~Yan, D.~J. Cordato, N.~W. Manning, M.~W. Parsons, R.~A. Hanel, A.~N.
  Aghaebrahim, T.~Y. Wu, P.~Cardona-Portela, N.~P\'{e}rez de~la Ossa, J.~D.
  Schaafsma, J.~Blasco, N.~Sangha, S.~Warach, C.~D. Gandhi, T.~J. Kleinig,
  D.~Sahlein, L.~Elijovich, W.~Tekle, E.~A. Samaniego, L.~Maali, M.~A.
  Abdulrazzak, M.~N. Psychogios, A.~Shuaib, D.~K. Pujara, F.~Shaker, H.~Johns,
  G.~Sharma, V.~Yogendrakumar, F.~C. Ng, M.~H. Rahbar, C.~Cai, P.~Lavori,
  S.~Hamilton, T.~Nguyen, J.~T. Fifi, S.~Davis, L.~Wechsler, V.~M. Pereira,
  M.~G. Lansberg, M.~D. Hill, J.~C. Grotta, M.~Ribo, B.~C. Campbell, and G.~W.
  Albers, ``Trial of endovascular thrombectomy for large ischemic strokes,''
  \emph{New England Journal of Medicine}. [Online]. Available:
  \url{https://doi.org/10.1056/NEJMoa2214403}
\BIBentrySTDinterwordspacing

\bibitem{Nogueira6to24}
\BIBentryALTinterwordspacing
R.~G. Nogueira, A.~P. Jadhav, D.~C. Haussen, A.~Bonafe, R.~F. Budzik, P.~Bhuva,
  D.~R. Yavagal, M.~Ribo, C.~Cognard, R.~A. Hanel, C.~A. Sila, A.~E. Hassan,
  M.~Millan, E.~I. Levy, P.~Mitchell, M.~Chen, J.~D. English, Q.~A. Shah, F.~L.
  Silver, V.~M. Pereira, B.~P. Mehta, B.~W. Baxter, M.~G. Abraham, P.~Cardona,
  E.~Veznedaroglu, F.~R. Hellinger, L.~Feng, J.~F. Kirmani, D.~K. Lopes, B.~T.
  Jankowitz, M.~R. Frankel, V.~Costalat, N.~A. Vora, A.~J. Yoo, A.~M. Malik,
  A.~J. Furlan, M.~Rubiera, A.~Aghaebrahim, J.-M. Olivot, W.~G. Tekle,
  R.~Shields, T.~Graves, R.~J. Lewis, W.~S. Smith, D.~S. Liebeskind, J.~L.
  Saver, and T.~G. Jovin, ``Thrombectomy 6 to 24 hours after stroke with a
  mismatch between deficit and infarct,'' \emph{New England Journal of
  Medicine}, vol. 378, no.~1, pp. 11--21, 2017. [Online]. Available:
  \url{https://www.nejm.org/doi/full/10.1056/NEJMoa1706442}
\BIBentrySTDinterwordspacing

\bibitem{hacke2008thrombolysis}
W.~Hacke, M.~Kaste, E.~Bluhmki, M.~Brozman, A.~D{\'a}valos, D.~Guidetti,
  V.~Larrue, K.~R. Lees, Z.~Medeghri, T.~Machnig \emph{et~al.}, ``Thrombolysis
  with alteplase 3 to 4.5 hours after acute ischemic stroke,'' \emph{New
  England journal of medicine}, vol. 359, no.~13, pp. 1317--1329, 2008.

\bibitem{KimUtilization}
\BIBentryALTinterwordspacing
Y.~Kim, S.~Lee, R.~Abdelkhaleq, V.~Lopez-Rivera, B.~Navi, H.~Kamel, S.~I.
  Savitz, A.~L. Czap, J.~C. Grotta, L.~D. McCullough, T.~M. Krause,
  L.~Giancardo, F.~S. Vahidy, and S.~A. Sheth, ``Utilization and availability
  of advanced imaging in patients with acute ischemic stroke,'' \emph{Circ
  Cardiovasc Qual Outcomes}, vol.~14, no.~4, p. e006989, 2021, kim, Youngran
  Lee, Songmi Abdelkhaleq, Rania Lopez-Rivera, Victor Navi, Babak Kamel, Hooman
  Savitz, Sean I Czap, Alexandra L Grotta, James C McCullough, Louise D Krause,
  Trudy Millard Giancardo, Luca Vahidy, Farhaan S Sheth, Sunil A eng Research
  Support, Non-U.S. Gov't 2021/03/25 Circ Cardiovasc Qual Outcomes. 2021
  Apr;14(4):e006989. doi: 10.1161/CIRCOUTCOMES.120.006989. Epub 2021 Mar 24.
  [Online]. Available: \url{https://www.ncbi.nlm.nih.gov/pubmed/33757311
  https://www.ahajournals.org/doi/pdf/10.1161/CIRCOUTCOMES.120.006989?download=true}
\BIBentrySTDinterwordspacing

\bibitem{McDonoughState}
\BIBentryALTinterwordspacing
R.~McDonough, J.~Ospel, and M.~Goyal, ``State of the art stroke imaging: A
  current perspective,'' \emph{Can Assoc Radiol J}, p. 8465371211028823, 2021,
  mcDonough, Rosalie Ospel, Johanna Goyal, Mayank eng 2021/09/28 Can Assoc
  Radiol J. 2021 Sep 27:8465371211028823. doi: 10.1177/08465371211028823.
  [Online]. Available: \url{https://www.ncbi.nlm.nih.gov/pubmed/34569306
  https://journals.sagepub.com/doi/pdf/10.1177/08465371211028823}
\BIBentrySTDinterwordspacing

\bibitem{Albers6to16}
\BIBentryALTinterwordspacing
G.~W. Albers, M.~P. Marks, S.~Kemp, S.~Christensen, J.~P. Tsai,
  S.~Ortega-Gutierrez, R.~A. McTaggart, M.~T. Torbey, M.~Kim-Tenser,
  T.~Leslie-Mazwi, A.~Sarraj, S.~E. Kasner, S.~A. Ansari, S.~D. Yeatts,
  S.~Hamilton, M.~Mlynash, J.~J. Heit, G.~Zaharchuk, S.~Kim, J.~Carrozzella,
  Y.~Y. Palesch, A.~M. Demchuk, R.~Bammer, P.~W. Lavori, J.~P. Broderick, and
  M.~G. Lansberg, ``Thrombectomy for stroke at 6 to 16 hours with selection by
  perfusion imaging,'' \emph{New England Journal of Medicine}, vol. 378, no.~8,
  pp. 708--718, 2018. [Online]. Available:
  \url{https://www.nejm.org/doi/full/10.1056/NEJMoa1713973
  https://www.nejm.org/doi/pdf/10.1056/NEJMoa1713973?articleTools=true}
\BIBentrySTDinterwordspacing

\bibitem{WangSocioeconomic}
\BIBentryALTinterwordspacing
J.~J. Wang, A.~Boltyenkov, J.~M. Katz, J.~O’Hara, M.~Gribko, and P.~C.
  Sanelli, ``Striving for socioeconomic equity in ischemic stroke care: Imaging
  and acute treatment utilization from a comprehensive stroke center,''
  \emph{Journal of the American College of Radiology}, vol.~19, no. 2, Part B,
  pp. 348--358, 2022. [Online]. Available:
  \url{https://www.sciencedirect.com/science/article/pii/S1546144021007341}
\BIBentrySTDinterwordspacing

\bibitem{phan2006aspects}
T.~G. Phan, G.~A. Donnan, M.~Koga, L.~A. Mitchell, M.~Molan, G.~Fitt, W.~Chong,
  M.~Holt, and D.~C. Reutens, ``The aspects template is weighted in favor of
  the striatocapsular region,'' \emph{Neuroimage}, vol.~31, no.~2, pp.
  477--481, 2006.

\bibitem{SchroederCritical}
\BIBentryALTinterwordspacing
J.~Schroeder and G.~Thomalla, ``A critical review of alberta stroke program
  early ct score for evaluation of acute stroke imaging,'' \emph{Frontiers in
  Neurology}, vol.~7, 2017. [Online]. Available:
  \url{https://www.frontiersin.org/article/10.3389/fneur.2016.00245}
\BIBentrySTDinterwordspacing

\bibitem{weaver2021strategic}
N.~A. Weaver, H.~J. Kuijf, H.~P. Aben, J.~Abrigo, H.-J. Bae, M.~Barbay, J.~G.
  Best, R.~Bordet, F.~M. Chappell, C.~P. Chen \emph{et~al.}, ``Strategic
  infarct locations for post-stroke cognitive impairment: a pooled analysis of
  individual patient data from 12 acute ischaemic stroke cohorts,'' \emph{The
  Lancet Neurology}, vol.~20, no.~6, pp. 448--459, 2021.

\bibitem{wang2019stroke}
Y.~Wang, J.~M. Juliano, S.-L. Liew, A.~M. McKinney, and S.~Payabvash, ``Stroke
  atlas of the brain: Voxel-wise density-based clustering of infarct lesions
  topographic distribution,'' \emph{NeuroImage: Clinical}, vol.~24, p. 101981,
  2019.

\bibitem{meyer2016voxel}
S.~Meyer, S.~S. Kessner, B.~Cheng, M.~B{\"o}nstrup, R.~Schulz, F.~C. Hummel,
  N.~De~Bruyn, A.~Peeters, V.~Van~Pesch, T.~Duprez \emph{et~al.}, ``Voxel-based
  lesion-symptom mapping of stroke lesions underlying somatosensory deficits,''
  \emph{NeuroImage: Clinical}, vol.~10, pp. 257--266, 2016.

\bibitem{broocks2018computed}
G.~Broocks, F.~Flottmann, M.~Ernst, T.~D. Faizy, J.~Minnerup, S.~Siemonsen,
  J.~Fiehler, and A.~Kemmling, ``Computed tomography--based imaging of
  voxel-wise lesion water uptake in ischemic brain: relationship between
  density and direct volumetry,'' \emph{Investigative radiology}, vol.~53,
  no.~4, pp. 207--213, 2018.

\bibitem{nowinski2020ischemic}
W.~L. Nowinski, J.~Walecki, G.~P{\'o}{\l}torak-Szymczak, K.~Sklinda, and
  B.~Mruk, ``Ischemic infarct detection, localization, and segmentation in
  noncontrast ct human brain scans: review of automated methods,''
  \emph{PeerJ}, vol.~8, p. e10444, 2020.

\bibitem{chen2022prognosis}
X.~Chen, S.~Lin, X.~Zhang, S.~Hu, and X.~Wang, ``Prognosis with non-contrast ct
  and ct perfusion imaging in thrombolysis-treated acute ischemic stroke,''
  \emph{European Journal of Radiology}, vol. 149, p. 110217, 2022.

\bibitem{el2022evaluating}
H.~El-Hariri, L.~A. S.~M. Neto, P.~Cimflova, F.~Bala, R.~Golan, A.~Sojoudi,
  C.~Duszynski, I.~Elebute, S.~H. Mousavi, W.~Qiu \emph{et~al.}, ``Evaluating
  nnu-net for early ischemic change segmentation on non-contrast computed
  tomography in patients with acute ischemic stroke,'' \emph{Computers in
  biology and medicine}, vol. 141, p. 105033, 2022.

\bibitem{LiuDetection}
\BIBentryALTinterwordspacing
C.-F. Liu, J.~Hsu, X.~Xu, S.~Ramachandran, V.~Wang, M.~I. Miller, A.~E. Hillis,
  A.~V. Faria, M.~Wintermark, S.~J. Warach, G.~W. Albers, S.~M. Davis, J.~C.
  Grotta, W.~Hacke, D.-W. Kang, C.~Kidwell, W.~J. Koroshetz, K.~R. Lees, M.~H.
  Lev, D.~S. Liebeskind, A.~G. Sorensen, V.~N. Thijs, G.~Thomalla, J.~M.
  Wardlaw, M.~Luby, S.~The, and V.~I. investigators, ``Deep learning-based
  detection and segmentation of diffusion abnormalities in acute ischemic
  stroke,'' \emph{Communications Medicine}, vol.~1, no.~1, p.~61, 2021.
  [Online]. Available: \url{https://doi.org/10.1038/s43856-021-00062-8
  https://www.nature.com/articles/s43856-021-00062-8.pdf}
\BIBentrySTDinterwordspacing

\bibitem{yu2020use}
Y.~Yu, Y.~Xie, T.~Thamm, E.~Gong, J.~Ouyang, C.~Huang, S.~Christensen, M.~P.
  Marks, M.~G. Lansberg, G.~W. Albers \emph{et~al.}, ``Use of deep learning to
  predict final ischemic stroke lesions from initial magnetic resonance
  imaging,'' \emph{JAMA network open}, vol.~3, no.~3, pp. e200\,772--e200\,772,
  2020.

\bibitem{warfield2004simultaneous}
S.~K. Warfield, K.~H. Zou, and W.~M. Wells, ``Simultaneous truth and
  performance level estimation (staple): an algorithm for the validation of
  image segmentation,'' \emph{IEEE transactions on medical imaging}, vol.~23,
  no.~7, pp. 903--921, 2004.

\bibitem{saver2006time}
J.~L. Saver, ``Time is brain—quantified,'' \emph{Stroke}, vol.~37, no.~1, pp.
  263--266, 2006.

\bibitem{marks2014effect}
M.~P. Marks, M.~G. Lansberg, M.~Mlynash, J.-M. Olivot, M.~Straka, S.~Kemp,
  R.~McTaggart, M.~Inoue, G.~Zaharchuk, R.~Bammer \emph{et~al.}, ``Effect of
  collateral blood flow on patients undergoing endovascular therapy for acute
  ischemic stroke,'' \emph{Stroke}, vol.~45, no.~4, pp. 1035--1039, 2014.

\bibitem{johnston2000predictive}
K.~Johnston, A.~Connors~Jr, D.~Wagner, W.~Knaus, X.-Q. Wang, and E.~C.
  Haley~Jr, ``A predictive risk model for outcomes of ischemic stroke,''
  \emph{Stroke}, vol.~31, no.~2, pp. 448--455, 2000.

\bibitem{ChristensenOptimizing}
\BIBentryALTinterwordspacing
S.~Christensen, M.~Mlynash, J.~MacLaren, C.~Federau, G.~W. Albers, and M.~G.
  Lansberg, ``Optimizing deep learning algorithms for segmentation of acute
  infarcts on non-contrast material-enhanced ct scans of the brain using
  simulated lesions,'' \emph{Radiol Artif Intell}, vol.~3, no.~4, p. e200127,
  2021, christensen, Soren Mlynash, Michael MacLaren, Julian Federau, Christian
  Albers, Gregory W Lansberg, Maarten G eng 2021/08/06 Radiol Artif Intell.
  2021 May 12;3(4):e200127. doi: 10.1148/ryai.2021200127. eCollection 2021 Jul.
  [Online]. Available: \url{https://www.ncbi.nlm.nih.gov/pubmed/34350404
  https://www.ncbi.nlm.nih.gov/pmc/articles/PMC8328101/pdf/ryai.2021200127.pdf}
\BIBentrySTDinterwordspacing

\bibitem{IsenseennU-Net}
\BIBentryALTinterwordspacing
F.~Isensee, P.~F. Jaeger, S.~A.~A. Kohl, J.~Petersen, and K.~H. Maier-Hein,
  ``nnu-net: a self-configuring method for deep learning-based biomedical image
  segmentation,'' \emph{Nat Methods}, vol.~18, no.~2, pp. 203--211, 2021,
  isensee, Fabian Jaeger, Paul F Kohl, Simon A A Petersen, Jens Maier-Hein,
  Klaus H eng Research Support, Non-U.S. Gov't 2020/12/09 Nat Methods. 2021
  Feb;18(2):203-211. doi: 10.1038/s41592-020-01008-z. Epub 2020 Dec 7.
  [Online]. Available: \url{https://www.ncbi.nlm.nih.gov/pubmed/33288961
  https://www.nature.com/articles/s41592-020-01008-z.pdf}
\BIBentrySTDinterwordspacing

\bibitem{cawley2010over}
G.~C. Cawley and N.~L. Talbot, ``On over-fitting in model selection and
  subsequent selection bias in performance evaluation,'' \emph{The Journal of
  Machine Learning Research}, vol.~11, pp. 2079--2107, 2010.

\bibitem{lakens2017equivalence}
D.~Lakens, ``Equivalence tests: A practical primer for t tests, correlations,
  and meta-analyses,'' \emph{Social psychological and personality science},
  vol.~8, no.~4, pp. 355--362, 2017.

\bibitem{ostmeier2022evaluation}
S.~Ostmeier, B.~Axelrod, J.~Bertels, F.~Isensee, M.~G. Lansberg,
  S.~Christensen, G.~W. Albers, L.-J. Li, and J.~J. Heit, ``Evaluation of
  medical image segmentation models for uncertain, small or empty reference
  annotations,'' \emph{arXiv preprint arXiv:2209.13008}, 2022.

\bibitem{nikolov2018deep}
S.~Nikolov, S.~Blackwell, A.~Zverovitch, R.~Mendes, M.~Livne, J.~De~Fauw,
  Y.~Patel, C.~Meyer, H.~Askham, B.~Romera-Paredes \emph{et~al.}, ``Deep
  learning to achieve clinically applicable segmentation of head and neck
  anatomy for radiotherapy,'' \emph{arXiv preprint arXiv:1809.04430}, 2018.

\bibitem{Maier_Hein_pitfall_metric}
L.~Maier-Hein, A.~Reinke, E.~Christodoulou, B.~Glocker, P.~Godau, F.~Isensee,
  J.~Kleesiek, M.~Kozubek, M.~Reyes, and M.~A. Riegler, ``Metrics reloaded:
  Pitfalls and recommendations for image analysis validation,'' \emph{arXiv
  preprint arXiv:2206.01653}, 2022.

\bibitem{vincent2021impact}
O.~Vincent, C.~Gros, and J.~Cohen-Adad, ``Impact of individual rater style on
  deep learning uncertainty in medical imaging segmentation,'' \emph{arXiv
  preprint arXiv:2105.02197}, 2021.

\bibitem{Langerak2008LabelFusion}
T.~R. Langerak, U.~A. van~der Heide, A.~N. T.~J. Kotte, M.~A. Viergever, M.~van
  Vulpen, and J.~P.~W. Pluim, ``Label fusion in atlas-based segmentation using
  a selective and iterative method for performance level estimation (simple),''
  \emph{IEEE Transactions on Medical Imaging}, vol.~29, no.~12, pp. 2000--2008,
  2010.

\bibitem{Landman2012StatisticalFusion}
B.~A. Landman, A.~J. Asman, A.~G. Scoggins, J.~A. Bogovic, F.~Xing, and J.~L.
  Prince, ``Robust statistical fusion of image labels,'' \emph{IEEE
  Transactions on Medical Imaging}, vol.~31, no.~2, pp. 512--522, 2012.

\bibitem{goyal2015randomized}
M.~Goyal, A.~M. Demchuk, B.~K. Menon, M.~Eesa, J.~L. Rempel, J.~Thornton,
  D.~Roy, T.~G. Jovin, R.~A. Willinsky, B.~L. Sapkota \emph{et~al.},
  ``Randomized assessment of rapid endovascular treatment of ischemic stroke,''
  \emph{New England Journal of Medicine}, vol. 372, no.~11, pp. 1019--1030,
  2015.

\bibitem{hacke1995intravenous}
W.~Hacke, M.~Kaste, C.~Fieschi, D.~Toni, E.~Lesaffre, R.~Von~Kummer, G.~Boysen,
  E.~Bluhmki, G.~H{\"o}xter, M.-H. Mahagne \emph{et~al.}, ``Intravenous
  thrombolysis with recombinant tissue plasminogen activator for acute
  hemispheric stroke: the european cooperative acute stroke study (ecass),''
  \emph{Jama}, vol. 274, no.~13, pp. 1017--1025, 1995.

\bibitem{zha2022non}
A.~M. Zha, H.~Kamal, J.~A. Jeevarajan, O.~Arevalo, L.~Zhu, C.~M. Ankrom, E.~E.
  Bonfante-Mejia, T.~D. Cossey, T.~C. Wu, A.~D. Barreto \emph{et~al.},
  ``Non-contrast head ct-based thrombolysis for wake-up/unknown onset stroke is
  safe: A single-center study and meta-analysis,'' \emph{International Journal
  of Stroke}, vol.~17, no.~3, pp. 354--361, 2022.

\bibitem{hill2014alberta}
M.~D. Hill, A.~M. Demchuk, M.~Goyal, T.~G. Jovin, L.~D. Foster, T.~A. Tomsick,
  R.~von Kummer, S.~D. Yeatts, Y.~Y. Palesch, and J.~P. Broderick, ``Alberta
  stroke program early computed tomography score to select patients for
  endovascular treatment: Interventional management of stroke (ims)-iii
  trial,'' \emph{Stroke}, vol.~45, no.~2, pp. 444--449, 2014.

\bibitem{yoo2014impact}
A.~J. Yoo, O.~O. Zaidat, Z.~A. Chaudhry, O.~A. Berkhemer, R.~G. Gonz{\'a}lez,
  M.~Goyal, A.~M. Demchuk, B.~K. Menon, E.~Mualem, D.~Ueda \emph{et~al.},
  ``Impact of pretreatment noncontrast ct alberta stroke program early ct score
  on clinical outcome after intra-arterial stroke therapy,'' \emph{Stroke},
  vol.~45, no.~3, pp. 746--751, 2014.

\bibitem{barber2000validity}
P.~A. Barber, A.~M. Demchuk, J.~Zhang, and A.~M. Buchan, ``Validity and
  reliability of a quantitative computed tomography score in predicting outcome
  of hyperacute stroke before thrombolytic therapy,'' \emph{The Lancet}, vol.
  355, no. 9216, pp. 1670--1674, 2000.

\bibitem{GoyalEndovascular}
\BIBentryALTinterwordspacing
M.~Goyal, B.~K. Menon, W.~H. van Zwam, D.~W. Dippel, P.~J. Mitchell, A.~M.
  Demchuk, A.~Dávalos, C.~B. Majoie, A.~van~der Lugt, and M.~A. De~Miquel,
  ``Endovascular thrombectomy after large-vessel ischaemic stroke: a
  meta-analysis of individual patient data from five randomised trials,''
  \emph{The Lancet}, vol. 387, no. 10029, pp. 1723--1731, 2016. [Online].
  Available:
  \url{https://www.sciencedirect.com/science/article/pii/S014067361600163X?via%3Dihub}
\BIBentrySTDinterwordspacing

\bibitem{BarberValidity}
\BIBentryALTinterwordspacing
P.~A. Barber, A.~M. Demchuk, J.~Zhang, and A.~M. Buchan, ``Validity and
  reliability of a quantitative computed tomography score in predicting outcome
  of hyperacute stroke before thrombolytic therapy,'' \emph{The Lancet}, vol.
  355, no. 9216, pp. 1670--1674, 2000, doi: 10.1016/S0140-6736(00)02237-6.
  [Online]. Available: \url{https://doi.org/10.1016/S0140-6736(00)02237-6
  https://www.sciencedirect.com/science/article/pii/S0140673600022376?via%3Dihub}
\BIBentrySTDinterwordspacing

\bibitem{sarraj2022accuracy}
A.~Sarraj, B.~C. Campbell, S.~Christensen, C.~W. Sitton, S.~Khanpara, R.~F.
  Riascos, D.~Pujara, F.~Shaker, G.~Sharma, M.~G. Lansberg \emph{et~al.},
  ``Accuracy of ct perfusion--based core estimation of follow-up infarction:
  effects of time since last known well,'' \emph{Neurology}, vol.~98, no.~21,
  pp. e2084--e2096, 2022.

\bibitem{KuangSegASPECT}
\BIBentryALTinterwordspacing
H.~Kuang, B.~K. Menon, S.~I.~L. Sohn, and W.~Qiu, ``Eis-net: Segmenting early
  infarct and scoring aspects simultaneously on non-contrast ct of patients
  with acute ischemic stroke,'' \emph{Medical Image Analysis}, vol.~70, p.
  101984, 2021. [Online]. Available:
  \url{https://www.sciencedirect.com/science/article/pii/S136184152100030X}
\BIBentrySTDinterwordspacing

\bibitem{QiuNCCTMachine}
\BIBentryALTinterwordspacing
W.~Qiu, H.~Kuang, E.~Teleg, J.~M. Ospel, S.~I. Sohn, M.~Almekhlafi, M.~Goyal,
  M.~D. Hill, A.~M. Demchuk, and B.~K. Menon, ``Machine learning for detecting
  early infarction in acute stroke with non-contrast-enhanced ct,''
  \emph{Radiology}, vol. 294, no.~3, pp. 638--644, 2020, qiu, Wu Kuang, Hulin
  Teleg, Ericka Ospel, Johanna M Sohn, Sung Il Almekhlafi, Mohammed Goyal,
  Mayank Hill, Michael D Demchuk, Andrew M Menon, Bijoy K eng CIHR/Canada
  Research Support, Non-U.S. Gov't 2020/01/29 Radiology. 2020
  Mar;294(3):638-644. doi: 10.1148/radiol.2020191193. Epub 2020 Jan 28.
  [Online]. Available: \url{https://www.ncbi.nlm.nih.gov/pubmed/31990267
  https://pubs.rsna.org/doi/pdf/10.1148/radiol.2020191193}
\BIBentrySTDinterwordspacing

\bibitem{gauriau2023head}
R.~Gauriau, B.~C. Bizzo, D.~S. Comeau, J.~M. Hillis, C.~P. Bridge, J.~K. Chin,
  J.~Pawar, A.~Pourvaziri, I.~Sesic, E.~Sharaf \emph{et~al.}, ``Head ct deep
  learning model is highly accurate for early infarct estimation,''
  \emph{Scientific Reports}, vol.~13, no.~1, p. 189, 2023.

\bibitem{lu2022identification}
J.~Lu, Y.~Zhou, W.~Lv, H.~Zhu, T.~Tian, S.~Yan, Y.~Xie, D.~Wu, Y.~Li, Y.~Liu
  \emph{et~al.}, ``Identification of early invisible acute ischemic stroke in
  non-contrast computed tomography using two-stage deep-learning model,''
  \emph{Theranostics}, vol.~12, no.~12, p. 5564, 2022.

\bibitem{albers1998diffusion}
G.~W. Albers, ``Diffusion-weighted mri for evaluation of acute stroke,''
  \emph{Neurology}, vol.~51, no. 3 Suppl 3, pp. S47--S49, 1998.

\bibitem{almandoz2011imaging}
J.~E.~D. Almandoz, S.~R. Pomerantz, R.~G. Gonz{\'a}lez, and M.~H. Lev,
  ``Imaging of acute ischemic stroke: Unenhanced computed tomography,''
  \emph{Acute Ischemic Stroke}, pp. 43--56, 2011.

\bibitem{lansberg2000comparison}
M.~G. Lansberg, G.~W. Albers, C.~Beaulieu, and M.~P. Marks, ``Comparison of
  diffusion-weighted mri and ct in acute stroke,'' \emph{Neurology}, vol.~54,
  no.~8, pp. 1557--1561, 2000.

\bibitem{christensenCostDriver}
E.~W. Christensen, C.~E. Pelzl, J.~Hemingway, J.~J. Wang, M.~X. Sanmartin,
  J.~J. Naidich, E.~Y. Rula, and P.~C. Sanelli, ``Drivers of ischemic stroke
  hospital cost trends among older adults in the united states,'' \emph{Journal
  of the American College of Radiology}, 2022.

\bibitem{Straka_time}
\BIBentryALTinterwordspacing
M.~Straka, G.~W. Albers, and R.~Bammer, ``Real-time diffusion-perfusion
  mismatch analysis in acute stroke,'' \emph{Journal of Magnetic Resonance
  Imaging}, vol.~32, no.~5, pp. 1024--1037, 2010. [Online]. Available:
  \url{https://onlinelibrary.wiley.com/doi/abs/10.1002/jmri.22338}
\BIBentrySTDinterwordspacing

\bibitem{kunz_monetary_benefit_2020}
W.~G. Kunz, M.~G. Hunink, M.~A. Almekhlafi, B.~K. Menon, J.~L. Saver, D.~W.
  Dippel, C.~B. Majoie, T.~G. Jovin, A.~Davalos, S.~Bracard, and et~al.,
  ``Public health and cost consequences of time delays to thrombectomy for
  acute ischemic stroke,'' \emph{Neurology}, vol.~95, no.~18, 2020.

\bibitem{nguyen_non_2022}
T.~N. Nguyen, M.~Abdalkader, S.~Nagel, M.~M. Qureshi, M.~Ribo, F.~Caparros,
  D.~C. Haussen, M.~H. Mohammaden, S.~A. Sheth, S.~Ortega-Gutierrez, and
  et~al., ``Noncontrast computed tomography vs computed tomography perfusion or
  magnetic resonance imaging selection in late presentation of stroke with
  large-vessel occlusion,'' \emph{JAMA Neurology}, vol.~79, no.~1, p.~22, 2022.

\end{thebibliography}
\pagebreak[4]

\FloatBarrier
\section{Tables}
% latex table generated in R 4.2.2 by xtable 1.8-4 package
% Fri Jan 27 13:19:55 2023

% latex table generated in R 4.2.2 by xtable 1.8-4 package
% Fri Jan 27 13:28:46 2023
\begin{table}[ht]
\caption{Patient characteristics}
\begin{NiceTabular}{|>{\bfseries}p{1.8cm}|>{\bfseries}p{4.8cm}|p{2cm}p{2cm}p{2cm}|p{2cm}|l|}
  \hline
\RowStyle{\bfseries} Categories & Characteristic & Randomized-Train$^1$ & Non-randomized-Train$^1$ & Total-Train$^1$ & Total-Test& p-value$^7$\\ 
  \hline
General & Total Number & 146 & 87 & 233 & 35 &  \\ 
   & Age & 70 (59-79) & 67 (58-76) & 69 (59-78) & 71 (64-78) & 0.4 \\ 
   & Female \% & 52 & 52 & 52 & 54 &  \\ 
  \multirow{3}{0pt}{Imaging Character-istics} & Expert A Volume [ml] & 9 (4-21) & 22 (6-69) & 12 (5-32) &  &  \\ 
   & Expert B Volume [ml] & 14 (5-27) & 14 (0-60) & 14 (4-37) &  &  \\ 
   & Expert C Volume [ml] & 3 (1-7) & 7 (0-43) & 4 (1-12) &  &  \\ 
   & Ischemic Core Volume [ml] & 9 (2-27) & 18 (0-77) & 11 (2-38) &  &  \\ 
   & Tmax6 Volume [ml]$^2$ & 117 (78-158) & 69 (3-150) & 104 (62-157) &  &  \\ 
   & ASPECTS on Baseline CT$^3$ & 8 (7-9) & 8 (5-10) & 8 (7-9) &  &  \\ 
  Process & Witnessed Number & 50 &  & 50 &  &  \\ 
   & Wake-Up Number & 77 &  & 77 &  &  \\ 
   & Unwitnessed Number & 19 &  & 19 &  &  \\ 
   & Onset to Image Time [h] & 10 (8-12) &  & 10 (8-12) & 10 (5-13) & 0.43 \\ 
\multirow{2}{0pt}{Follow-Up}  & 24h DWI Number & 146 &  & 146 &  &  \\ 
   & 24h DWI Volume [ml] & 39 (24-110) &  & 39 (24-110) &  &  \\ 
  \multirow{2}{0pt}{Clinical Outcome} & mRS$^5$ at Baseline & 0 (0-0) &  & 0 (0-0) & 0 (0-0) &  \\ 
   & mRS$^5$  at 90 days & 4 (2-5) &  & 4 (2-5) & 2 (1-3) & 0.002 \\ 
   \hline
\end{NiceTabular}
\footnotesize{\newline$^1$ Median (1st - 3rd quantile), if not otherwise indicated, $^2$ Time-to-Maximum after 6 seconds, $^3$ Alberta Stroke Program Early CT Score, $^5$ modified Ranking Scale, $^6$ data for non-randomized patients not available, $^7$ double sided Wilcoxon test}
\label{tab:pcharacteristics}
\end{table}

\begin{table*}
\caption{Segmentation, Internal Evaluation}
\label{tab:segmentation}
\begin{tabular}{|>{\bfseries}l>{\bfseries}l|rl|rll|rll|rl|}
\hline
\multirow{5}{0pt}{\textbf{Categories}} & \multirow{5}{0pt}{\textbf{Metric}} &  \multicolumn{8}{c|}{\textbf{Internal Evaluation}} & \multicolumn{2}{r|}{\textbf{External Evaluation}} \\   
 & & \multicolumn{2}{r}{\textbf{\begin{tabular}[l]{@{}l@{}}Random\\Sampling$^5$\end{tabular}}} & \multicolumn{2}{r}{\textbf{Interexpert$^5$}} & \multicolumn{1}{c}{\textbf{\begin{tabular}[c]{@{}l@{}}p-value \\ random-\\ inter-\\expert$^6$\end{tabular}}} & \multicolumn{2}{c}{\textbf{\begin{tabular}[c]{@{}l@{}}Majority\\Vote$^5$\end{tabular}}} &  \textbf{\begin{tabular}[c]{@{}l@{}}p-value \\ random-\\ majority$^6$\end{tabular}}& \multicolumn{2}{c|}{\textbf{\begin{tabular}[c]{@{}c@{}}Random\\Sampling$^7$\end{tabular}}}\\   \hline
Volume & VS & 0.71 & ± 0.04 & 0.55 & ± 0.06 & $<$0.0001 & 0.45 & ± 0.09 & $<$0.0001 & 0.49 & ±0.08\\ 
    & AVD [ml] & 5.36 & ± 0.87 & 8.85 & ± 1.46 & $<$0.0001 & 7.91 & ± 1.74 & $<$0.0001 & 5.47 & ±4.12\\ 
  Overlap & Dice & 0.51 & ± 0.04 & 0.36 & ± 0.05 & $<$0.0001 & 0.45 & ± 0.05 & $<$0.0001 &0.37 & ±0.07\\ 
    & Precision & 0.61 & ± 0.07 & 0.53 & ± 0.03 & $<$0.05 & 0.79 & ± 0.04 & non-sig  & 0.90 & ±0.06 \\ 
    & Recall & 0.60 & ± 0.05 & 0.47 & ± 0.03 & $<$0.0001 & 0.38 & ± 0.06 & $<$0.0001 & 0.24 & ±0.06 \\ 
  Distance & HD 95 [mm] & 13.60 & ± 2.01 & 17.54 & ± 2.01 & $<$0.0001 & 10.92 & ± 1.72 & non-sig & 18.11 & ±7.15 \\ 
    & SDT 5mm & 0.71 & ± 0.03 & 0.60 & ± 0.05 & $<$0.0001 & 0.75 & ± 0.03 & non-sig & 0.59 & ±0.07 \\ 
   \hline
\end{tabular}
\vspace{0.3\baselineskip}
\footnotesize{ \newline $^1$ VS = Volumetric Similarity,\newline$^2$ AVD = Absolute Volume Difference,\newline $^3$ HD 95 = Hausdorff Distance 95th percentile, \newline$^4$ SDT = Surface Dice at Tolerance, \newline $^5$ Median $\pm$ 95\% CI (bootstrapped) compared to Expert A, B, C, \newline$^6$ p-values of one-sided Wilcoxon sign rank test, \newline $^7$ Median $\pm$ 95\% CI (bootstrapped) compared to Expert D}
\end{table*}

%\begin{table}[]
% \begin{tabular}{lll}
% Dice & 0.36615204 & ±0.07 \\
% Hausdorff   Distance 95 & 18.113048 & ±7.15 \\
% Precision & 0.89894757 & ±0.06 \\
% Recall & 0.2433462 & ±0.06 \\
% Surface Dice   Variable 5 & 0.28837941 & ±0.07 \\
% Volume   Absolute Difference & 5.47248605 & ±4.12 \\
% Volumetric   Similarity & 0.48533783 & ±0.08
% \end{tabular}
% \end{table}

% latex table generated in R 4.2.2 by xtable 1.8-4 package
% Fri Jan 27 10:11:51 2023
\begin{table}[ht]
\caption{Image Classification with 1ml threshold}
\label{tab:classification}
\begin{tabular}{|>{\bfseries}l|>{\bfseries}l|rl|rl|rl|rl|}
  \hline
  \multirow{2}{0pt}{\textbf{Categories}} & \multirow{2}{0pt}{\textbf{Metric}} & \multicolumn{6}{c|}{\textbf{Internal Evaluation}} & \multicolumn{2}{r|}{\textbf{External Evaluation}} \\  
  & & \multicolumn{2}{r}{\textbf{Random Sampling$^1$}} & \multicolumn{2}{r}{\textbf{Interexpert$^1$}} & \multicolumn{2}{r|}{\textbf{Majority Vote$^1$}} & \multicolumn{2}{r|}{\textbf{Random Sampling $^2$}}\\
\hline
Image-level & Sensitivity & 0.94 & ± 0.02 & 0.91 & ± 0.02 & 0.65 & ± 0.03 & 0.95	& ±0.05\\ 
classification & Specificity & 0.70 & ± 0.03 & 0.99 & ± 0.03 & 0.97 & ± 0.01 & 0.36	& ±0.07\\ 
    & F-score & 0.85 & ± 0.02 & 0.85 & ± 0.02 & 0.77 & ± 0.02 &0.78 & ± 0.05 \\ 
    %& CCR  $^2$& 0.83 & ± 0.02 & 0.86 & ± 0.02 & 0.80 & ± 0.02 \\ 
    & AUC & 0.92 & ± 0.02 & 0.93 & ± 0.02 & 0.90 & ± 0.03 & 0.74 &	±0.09 \\ 
   \hline
\end{tabular}
\vspace{0.3\baselineskip}
\footnotesize{ \newline $^1$ Median $\pm$ 95\% CI (bootstrapped) compared to Expert A, B, C 
%\newline $^2$ CCR = Correct classification ratio 
\newline $^2$ Median $\pm$ 95\% CI (bootstrapped) compared to Expert D}
\end{table}

\begin{table}[ht]
\caption{Correlation to the clinical outcome for ASPECTS and volume estimates}
\label{tab:clinical_outcome}
\begin{tabular}{|>{\bfseries}l|p{2cm}l|p{2cm}l|}
  \hline
\bfseries Predictor & \bfseries Rho of mRS 30 days$^1$ & \bfseries p-value & \bfseries Rho of mRS 90 days $^1$& \bfseries p-value \\ 
  \hline
ASPECT & -0.18 & $<$0.05 & -0.19 & $<$0.05 \\ 
  Ischemic core volume of CTP & 0.16 & non-sig & 0.15 & non-sig \\ 
  Median expert volume & 0.17 & $<$0.05 & 0.16 & non-sig \\ 
  Volume of majority vote model & 0.13 & non-sig & 0.15 & non-sig \\ 
  Volume of random expert sampling model & 0.21 & $<$0.01 & 0.19 & $<$0.05 \\ 
   \hline
\end{tabular}
\vspace{0.3\baselineskip}
\footnotesize{ \newline $^1$ Spearman's Rho and p-value for correlation \newline $^2$ Alberta Stroke Program Early CT Score \newline $^3$ Computer Tomography Perfusion, Cerebral blood flow (CBF) $<$30\%}
\end{table}

\FloatBarrier
\pagebreak[4]
\section{Figures}

\FloatBarrier
\section{Supplementary Material}
\label{sec:supp_mat}

\subsection{Supplemental Methods\label{sec:sub}}
\beginsupplement

\subsubsection{Mathematical derivation: Cross Entropy Loss as Product of Bernoulli Maximum Likelihood estimation \label{sec:cross_entropy}}
While using majority vote preprocessing of data results in a regression problem, optimizing the cross entropy loss corresponds to a maximum likelihood estimation by recovering the probability distribution that is most likely to have produced the data. 

In fact, maximizing the cross entropy loss corresponds to maximizing the likelihood of producing the data when assuming each voxel labeling is independent. 

Under that assumption, the log-likelihood of getting a labeling $\{k_i\}$, $i$ indexing voxels can be transformed into the cross entropy loss with the following standard transformation:
\begin{align*}
\mathrm{log\_likelihood} &= \log\left (\prod\limits_{i \mid k_i = 1} p_i \prod\limits_{i \mid k_i = 0} (1 - p_i) \right ) \\
 &= \log\left (\prod\limits_{i \mid k_i = 1} p_i \prod\limits_{i \mid k_i = 0} (1 - p_i) \right ) \\
&= \sum\limits_{i \mid k_i = 1} \log \left (  p_i\right ) + \sum\limits_{i \mid k_i = 0} \log (1 - p_i)  \\
&= \sum\limits_{i \mid k_i = 1} k_i\log \left (  p_i\right ) + \sum\limits_{i \mid k_i = 0} (1 - k_i)\log (1 - p_i)\\
&= \sum\limits_{i} k_i\log \left (  p_i\right ) + (1 - k_i)\log (1 - p_i) 
\end{align*}

\pagebreak[4]
\FloatBarrier

\subsection{Tabels}
\begin{table}[H]
\centering
\caption{\textbf{Definitions of Performance Metrics for Medical Image Segmentation}}
\label{table:metric_definition}
\begin{tabular}{|p{2cm}|>{\bfseries}p{5cm}|l|l|}
\hline  
\textbf{Category}  & \textbf{Metric}   & \textbf{Abbreviation} & \textbf{Definition} \\ 
\hline 
\Tstrut \textbf{Volume}    & Volumetric Similarity    & VS   & \Bstrut  \belowbaseline[0pt]{$1 - \frac{\left | \left | V_p^1\right | - \left | V_{ra}^1\right |\right |}{ \left | V_p^1\right | + \left | V_{ra}^1\right | + \epsilon}$} \\
& Absolute Volume Difference    & AVD   &  \Bstrut $\frac 1 m \sum\limits_{i = 1}^m \left | V_{ra}^i - V_p^i\right |$  \\ \hline
\Tstrut\Bstrut \textbf{Overlap}   & Dice Similarity Coefficient   & Dice &  \belowbaseline[0pt]{$\frac{2\times TP}{2\times TP + FN + FP}$}\\
& Recall = Sensitivity& Recall    &   $\frac{TP}{TP + FN}$ \\
& Precision & Precision &   \abovebaseline[0pt]{$\frac{TP}{TP + FP}$}\Bstrut \\ \hline
\Tstrut\Bstrut \textbf{Distance}  & Hausdorff Distance, q = 95th percentile\Tstrut\Bstrut & HD 95 & \belowbaseline[0pt]{$ \begin{aligned}    \max&\left (h(A,B), h(B,A) \right ) \textrm{ with } \\ & h(A,B) = \max\limits_{a \in A} \min\limits_{b \in B} || b - a || \end{aligned}$}\Tstrut\Bstrut\\
%& Average Surface Distance\Tstrut\Bstrut & ASD  &   \belowbaseline[0pt]{$\frac{1}{|S_{ra}| + |S_p|} \times \sum\limits_{s \in S_p} d \left (x,S_{ra} \right ) + \sum\limits_{y \in S_{ra}} d(y, S_p)$}\Tstrut\Bstrut\\
& Surface Dice at Tolerance& SDT  &   \belowbaseline[0pt]{$\frac{|S_p \cap B_{ra}^t| + |S_{ra} \cap B_p^t |}{|S_p| + |S_{ra}|}$}\Tstrut\Bstrut \\ \hline
\Tstrut \textbf{Image-level classification}
& Correct Classification Rate   & CCR  &  \belowbaseline[0pt]{$\frac{\textrm{number of correctly detected subjects}}{\textrm{number of all subjects}}$} \\
& Sensitivity   &  Sensitivity & $\frac{TP_i}{TP_i + FN_i}$ \\
& Specificity  & Specificity  &   $\frac{TN_i}{TN_i + FP_i}$ \\
& Area Under the Curve  & AUC  &  \\\hline
\end{tabular}
\end{table}

\begin{table}[]
\caption{\bfseries Segmentations precision}
\label{tab:seg_precision}
\begin{tabular}{|l|l|r|rrrl|rrrl|}
\hline
\textbf{Categories} & \textbf{Metric$^1$} & \multicolumn{1}{c|}{\textbf{\begin{tabular}[c]{@{}l@{}}Random\\ expert \\ sampling \\ variance\end{tabular}}} & \multicolumn{1}{c}{\textbf{\begin{tabular}[c]{@{}l@{}}Inter-\\expert \\ variance\end{tabular}}} & \multicolumn{2}{c}{\textbf{Ratio$^2$}} & \textbf{p-value} & \multicolumn{1}{c}{\textbf{\begin{tabular}[c]{@{}l@{}}Majority \\vote \\ variance\end{tabular}}} & \multicolumn{2}{c}{\textbf{Ratio$^3$}} & \textbf{p-value} \\ 
  \hline
Volume & VS & 0.05 & 0.06 & 0.91 & 0.23 & non-sig & 0.09 & 0.56 & 0.12 & $<$0.0001 \\ 
    & AVD [ml] & 263.38 & 418.18 & 0.63 & 0.25 & $<$0.05 & 487.67 & 0.54 & 0.19 & $<$0.0001 \\ 
  Overlap & Dice & 0.06 & 0.05 & 1.08 & 0.18 & non-sig & 0.06 & 0.90 & 0.13 & non-sig \\ 
    & Precision & 0.07 & 0.07 & 1.12 & 0.19 & non-sig & 0.10 & 0.72 & 0.09 & $<$0.0001 \\ 
    & Recall & 0.08 & 0.07 & 1.14 & 0.17 & non-sig & 0.06 & 1.37 & 0.23 & non-sig \\ 
  Distance & HD 95 [mm] & 306.87 & 159.76 & 1.92 & 0.86 & non-sig & 338.46 & 0.91 & 0.31 & non-sig \\ 
    & SDT 5mm & 0.06 & 0.05 & 1.22 & 0.29 & non-sig & 0.07 & 0.86 & 0.16 & non-sig \\ 
   \hline
\end{tabular}
\vspace{0.3\baselineskip}
\footnotesize{\newline $^1$ VS = Volumetric Similarity, AVD = Absolute Volume Difference, HD 95 = Hausdorff Distance 95th percentile, \\ SDT = Surface
Dice at Tolerance \newline $^2$ Ratio of inter-expert variance over random expert sampling variance ± Standard deviation, \newline $^3$ Ratio of majority vote over variance over random expert sampling variance ± Standard deviation}
\end{table}

\FloatBarrier

\subsection{Figures}

\end{document}